\documentclass[letterpaper, 10 pt, conference]{ieeeconf}  % Comment this line out if you need a4paper

\IEEEoverridecommandlockouts                              % This command is only needed if 
\title{\LARGE \bf
PhysGraph: A Physics-aware 3D Scene Graph \\ 
for Perception and Reasoning}

\author{Haoyu Li$^{1}$, Aaron Thomas$^{1}$,
Shuyan Zhou$^{2}$, Xianyi Cheng$^{1}$ 
\thanks{$^{1}$Department of Mechanical Engineering and Materials Science,
Duke University, Durham, NC 27708, USA.}
\thanks{$^{2}$Department of Computer Science,
Duke University, Durham, NC 27708, USA.}
}

\usepackage{hyperref}
\usepackage{booktabs}
\usepackage{multirow}
\usepackage{amsmath}
\usepackage{amssymb}
\usepackage{cite}
\usepackage[table]{xcolor}
\usepackage{graphicx}
\definecolor{best}{RGB}{255,178,178}     % #bde6cd
\definecolor{second}{RGB}{255,217,178}   % #e5efbf
\definecolor{third}{RGB}{255,255,178}
\newcommand{\Best}[1]{\cellcolor{best}#1}
\newcommand{\Sec}[1]{\cellcolor{second}#1}
\newcommand{\Thi}[1]{\cellcolor{third}#1}

\begin{document}

\maketitle
\thispagestyle{empty}
\pagestyle{empty}

%%%%%%%%%%%%%%%%%%%%%%%%%%%%%%%%%%%%%%%%%%%%%%%%%%%%%%%%%%%%%%%%%%%%%%%%%%%%%%%%
\begin{abstract}
To perform a wide range of daily tasks, robots need to construct a 3D representation that is semantically rich, physically grounded, and structured enough to support task planning and affordance prediction. However, existing approaches primarily focus on semantic retrieval that often overlooks the physical and kinematic factors. Methods that attempt to model physical properties typically rely on narrow training sets or single-object modeling, limiting scalability and generalization across diverse object types. 
To address these challenges, we present \textbf{PhysGraph}, a framework that unifies symbolic reasoning with structured 3D geometry to model kinematic and physical properties in cluttered scenes. Given RGB-D observations, PhysGraph reconstructs object-centric 3D geometry and associates object instances across views. It then decomposes objects into functional parts and infers materials and articulations through visual reasoning. Evaluated on both synthetic and real-world datasets, \textbf{PhysGraph} achieves state-of-the-art results in semantic segmentation, multi-object mass estimation, and articulation prediction. With its simple yet effective design, PhysGraph produces physically consistent and semantically structured scene graphs, serving as a structured 3D representation for downstream tasks such as constraint-aware 3D affordance prediction and real-to-sim transfer, both of which are demonstrated in our experiments. See our project page at \url{https://phys-graph.github.io/}.
\end{abstract}
\section{Introduction}
\label{sec:intro}

\begin{figure}[t]
    \centering
    \includegraphics[width=\columnwidth]{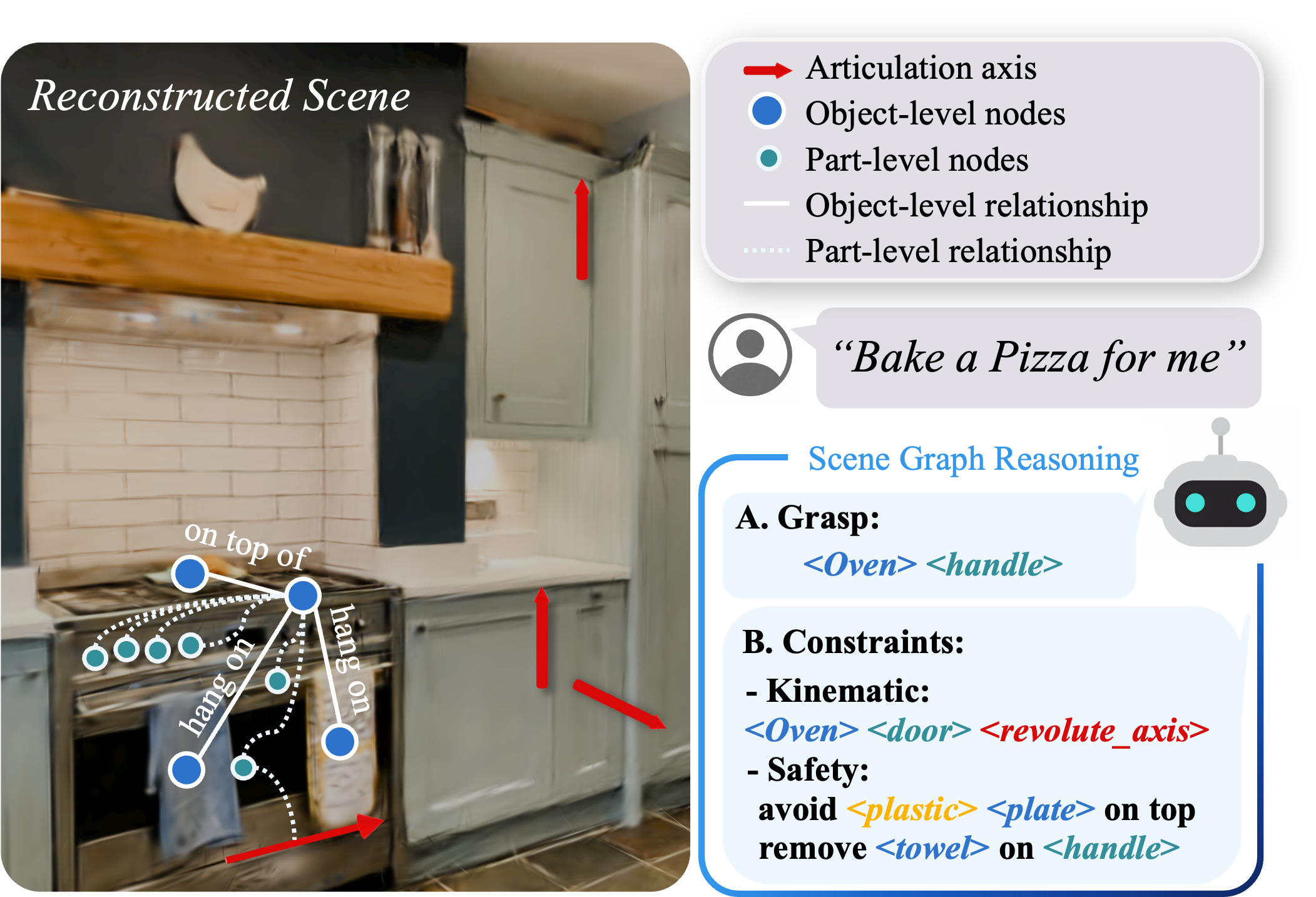}
    \caption{An example of generated hierarchical scene graph. \textbf{PhysGraph} builds a physics-grounded 3D scene graph that captures object-level spatial relations and part-level attributes, including articulation and material properties. This structured representation enables constraint-aware functional reasoning. Given the instruction ``Bake a pizza for me,'' the system grounds the 3D affordance of the oven handle, infers the door's revolute axis, and enforces safety constraints by avoiding heat-unsafe objects (e.g., plastic plates) and removing obstructing items (e.g., towels).}
    \label{fig:teaser}
\end{figure}

An informative scene representation can facilitate various robotic tasks, including task-driven active perception, path planning or long-horizon manipulation~\cite{irshad2024neuralfieldsroboticssurvey}. Building on the advance of Vision Foundation Models (VFMs), existing works~\cite{gu2024conceptgraphs,conceptfusion} enrich 3D representation with semantic information to support open-vocabulary query and zero-shot object retrieval, and they enable robots to understand and reason about their surroundings. However, semantic information alone is insufficient, robots also require an understanding of physical and kinematic information to perform precise interactions. As illustrated in Fig.~\ref{fig:teaser}: cooking utensils serve different purposes, plastic plates are unsuitable for oven use, and drawers exhibit distinct motion patterns depending on their hinge mechanisms. Such physical and kinematic knowledge imposes implicit constraints that directly affect task feasibility and safety, necessitating constraint-aware reasoning beyond semantic recognition. However, automatically inferring these constraints and integrating physical and kinematic attributes into existing 3D scene representations remains a significant and largely unresolved challenge.

To mitigate the domain gap in articulation estimation, a line of work~\cite{zhang2025iaao} focuses on modeling the motion of articulated objects and constructing interactable replicas that capture their kinematic behavior. These methods improve motion understanding but typically operate on isolated objects. They also rely on extensive human supervision to collect multiple configurations, which limits scalability. Another line of research~\cite{zhai2024physical, shuai2025pugs} estimates physical properties of individual objects under controlled settings. While effective in constrained environments, these approaches rarely generalize to complex real-world scenes with diverse objects and cluttered layout.  

To address these challenges, we propose PhysGraph, a framework that constructs a \textit{physics-grounded 3D scene representation} by unifying a hierarchical scene graph with inferred physical and kinematic attributes. We leverage visual foundation models~\cite{mobile_sam, siméoni2025dinov3} and vision language models~\cite{openai2025gpt5} to incrementally build a hierarchical scene graph that encodes object geometry, appearance, physical properties, spatial relationships, and kinematic constraints. This structured representation enables direct retrieval of task-relevant constraints for functional reasoning and safe manipulation. Our main contributions are as follows.
\begin{itemize}
    \item A constraint-aware hierarchical scene graph that supports task-driven 3D affordance grounding and constraint inference, facilitating precise articulation manipulation and long-horizon task planning.
    \item An effective LLM-based reasoning module that grounds physical knowledge in 3D space without training, enabling zero-shot articulation modeling across objects.
\end{itemize}

\begin{figure*}[t]
    \centering
    \includegraphics[width=\textwidth]{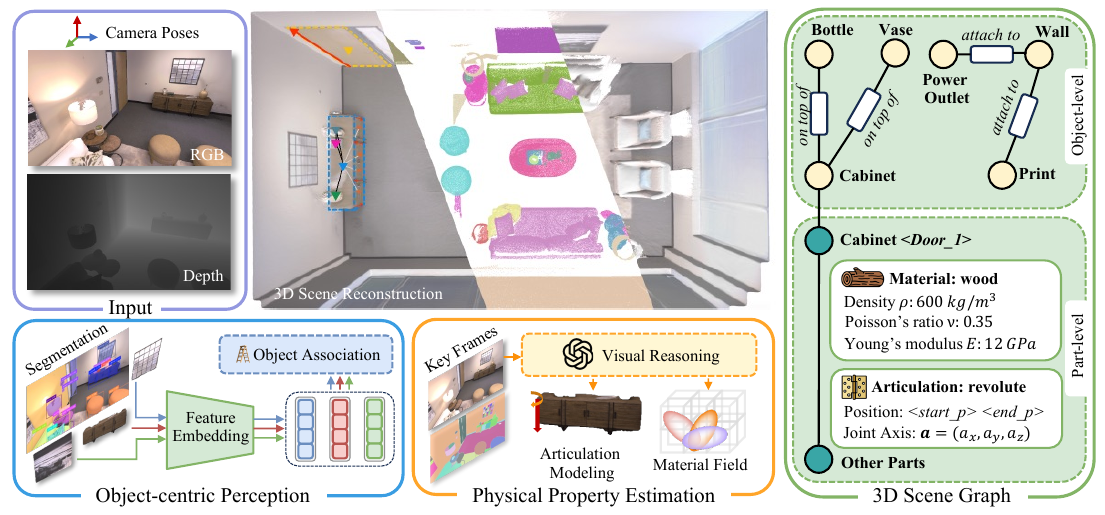}
    \caption{\textbf{Overview of PhysGraph.} 
    Given RGB-D observations and camera poses, PhysGraph first performs object-centric perception to reconstruct and segment objects, extract object-level features, and associate instances across views. Selected key frames are then used for visual reasoning to estimate part-level physical properties, including articulation and material attributes. The resulting outputs are integrated into a hierarchical 3D scene graph that encodes object relationships together with part-level physical properties.}
    \label{fig:overview}
    \vspace{-2mm}
\end{figure*}

\section{Related Work}
\label{sec:related_work}

\subsection{3D Scene Graph}

A 3D scene graph represents indoor environments as a graph of objects connected by spatial relations~\cite{koch2024lang3dsg, armeni20193d,rosinol20203d}. Functional 3D scene graphs extend this notion by adding interactive elements (e.g., switches, handles, buttons) as nodes and by modeling function-centric relations among objects and parts. IFR-Explore~\cite{li2021ifr} learns functional relationships via reinforcement learning. However, it is trained on closed-set dataset and lacks part-level reasoning, limiting its generalizability to complex scenes. Prior open-vocabulary scene graphs focus mainly on object-level and a narrow set of spatial predicates: Open3DSG~\cite{koch2024open3dsg} leverages object-level CLIP features~\cite{radford2021learning} yet struggles with part-level interactive elements and is largely constrained to proximity-based spatial edges, while ConceptGraph~\cite{radford2021learning} follows a direct inference pipeline but concentrates on object nodes and a small inventory of relations (e.g., \textit{on}, \textit{in}). OpenFunGraph~\cite{zhang2025open} infers functional relationships between objects but lacks reasoning about kinematics and physical dynamics, thereby limiting its applicability to robotics. In contrast, our method is open-vocabulary and supports part-level interactive elements.
It models both kinematic and physical relationships, producing a unified 3D representation for real-to-sim transfer and robotic task planning.

\subsection{Neural Fields for Robotic Manipulation}

Recently, the robotics community has shown growing interest in using neural fields to represent scene geometry and appearance~\cite{zhu2021rgb,shafiullah2022clip}. A major line of work~\cite{zhai2024physical, shen2023distilled, shuai2025pugs} distills neural
feature fields from visual foundation models~\cite{kirillov2023segment}.
NeRF2Physics~\cite{zhai2024physical} is the first to leverage large language models to infer
physical properties from NeRFs~\cite{mildenhall2021nerf}, but inherits NeRF’s costly
sampling procedure. PUGS~\cite{shuai2025pugs} performs physical reasoning on single objects,
making it unsuitable for multi-object scenes. Overall, these methods focus on individual
objects and require offline training, limiting their transferability to real-world robotic
applications. Other approaches use neural fields as scene representations for
manipulation~\cite{shen2023distilled}, yet they primarily
encode semantics and often overlook physical and kinematic constraints. In contrast, our method operates online and directly retrieves object-level and part-level semantics, articulation constraints, and material properties, yielding a physically grounded 3D scene representation.

\subsection{Functional and Affordance Understanding}
The ability to perceive, interpret, and interact with the environment is fundamental for
robotic agents and relies heavily on understanding object affordances within complex
scenes~\cite{hassanin2021visual}. Affordances describe the potential actions an environment
offers to an agent and are commonly inferred either through
probability maps~\cite{tang2025uad} or direct mappings from observations~\cite{huang2024rekep}. Probability-map–based approaches typically output
a distribution over the image space~\cite{mandikal2021learning}, but
these 2D representations lack geometric context and require depth data for real-world
deployment. Recent works~\cite{zhang2025iaao, yu2025artgs} instead adopt 3D representations
and multi-state rendering to model articulated objects by observing motion across
configurations. In contrast, our method introduces a generalizable reasoning module that directly estimates
multiple articulated objects along with their physical properties without additional human efforts, providing the structured
information necessary for more complex manipulation tasks.

\section{Method}

Given observations $\mathcal{O}=\{(I_k,D_k, T_k)\}_{k=0}^{K}$ of synchronized RGB $I_k$, depth images $D_k$ and camera extrinsics $T_k$, our goal is to recover a semantic rich and physics grounded 3D scene graph $\mathcal{G}_o$. We gradually detect and reconstruct objects in the scene, and associate them cross views by comparing their space position and semantic embeddings, which serve as the nodes $\mathcal{V}_o$ in our 3D scene graph. Then we leverage a LLM to infer spatial relationships between adjacent objects. We further decompose instances into functional parts $\mathcal{G}_p$, using visual prompting to retrieve physical parameter and articulation constraints, constituting part-level graph. This resultant hierarchical scene graph is open-vocabulary and constraint-aware, paving the way for semantic segmentation, articulation estimation, simulation-ready reconstruction and 3D affordance prediction. We show the overview in Fig.~\ref{fig:overview}.

\subsection{Object-centric Perception}
\label{p:o_perception}
Leveraging vision foundation models, we first detect object-level masks and extract corresponding visual and semantic features for each instance. We then reconstruct each object with an explicit Gaussian Splatting representation to model its geometry and appearance in a unified space. Based on the reconstructed geometry and aggregated embeddings, we associate object instances across multiple views using combined geometric and semantic similarity measures to maintain spatial and semantic consistency. 

\noindent \textbf{Object Representation}
To precisely modeling objects, we use 3DGS~\cite{kerbl20233d} as an explicit object representation $\mathbf{g}_i$ with high-fidelity appearance. Specifically, each 3D Gaussian is described by $\{\mathcal{X}, \boldsymbol{\Sigma}, \Phi\}$, 
where $\mathcal{X} \in \mathbb{R}^3$ is the centroid of the Gaussian, $\boldsymbol{\Sigma}$ is its covariance matrix in the world frame and $\phi$ is learned DINOv3~\cite{siméoni2025dinov3} feature vector. We can then render the depth map $\hat{\mathbf{D}}_k$, the color $\hat{\mathbf{C}}_k$, and the visual features $\hat{\mathbf{\Phi}}_k$ with the splatting algorithm:

\begin{equation}
\{\hat{\mathbf{\Phi}}_k, \hat{\mathbf{D}}_k,\hat{\mathbf{C}}_k\} 
= \sum_j \{\mathbf{\phi}_j, h_j, c_j\} \cdot \alpha_j 
\prod_{q=1}^{j-1} (1 - \alpha_q),
\end{equation}

where $\alpha_i$ is the opacity of the Gaussian conditioned on $\boldsymbol{\Sigma}'$, $h_j$ represent the depth of the Gaussian primitve and the indices $j \in N$ are in ascending order determined by their distance to the camera origin.

\noindent \textbf{Object Discovery}
Given observations $\{(I_k, D_k, T_k)\}_{k=0}^{K}$, we identify and associate objects across views to establish a consistent multi-view object set. This step ensures spatial and semantic coherence, providing a reliable object set for subsequent 3D scene graph generation and physical reasoning. For each frame $I_k$, we use YOLO-World~\cite{cheng2024yolo} and MobileSAMv2~\cite{mobile_sam} to produce object-level masks $\{M_o^i\} = \text{Seg}(I_k)$.
The image $I_k$ is then processed by a visual feature extractor to obtain global descriptors $\{F_C^k, F_D^k\} = \text{Embed}(I_k)$, which consist of CLIP features~\cite{radford2021learning} $F_C^k$ and DINOv3~\cite{siméoni2025dinov3} features $F_D^k$. For each detected object mask $M_o^i$ in frame $k$, we apply Masked Average Pooling(MAP) to aggregate object-level features $\mathbf{f}_k^i = \{\mathbf{f}_c^i, \mathbf{f}_d^i\}$:
\begin{equation}
\{\mathbf{f}_{c}^i,\mathbf{f}_{d}^i \} = MAP(M_o^i, F_C^i, F_D^i)=
\frac{\displaystyle \sum 
\mathbf{M}_o^i\,
\frac{\mathbf{F}_{C,D}^k}{\lVert \mathbf{F}_{C,D}^k \rVert}}
{\displaystyle \sum \mathbf{M}_o^i}
\end{equation}
This yields a semantic embedding for each object. To capture the spatial structure of each instance, we adopt 3D Gaussian Splatting~\cite{kerbl20233d} to model its geometry and appearance representation $\{\mathbf{g}_i, \mathbf{f}_i\}$. 
We then derive a compact spatial descriptor by constructing its 3D bounding box $B_i$, obtained by (i) filtering out low-opacity Gaussians with $\alpha < 0.1$, and (ii) computing an Axis-Aligned Bounding Box (AABB)~\cite{meister2021bvh} over the remaining points. This bounding box lays the foundation for object association and object voxelization. Each object instance is thus characterized by its bounding volume $B_i$ and the corresponding semantic feature $\mathbf{f}_k^i$

For each newly detected object $\{M_k^i, \mathbf{f}_k^i\}$, we evaluate its correspondence with existing instances by computing both spatial proximity $S^{\textrm{geo}}_{i,j}$ and semantic similarity $S^{\textrm{emb}}_{i,j}$ among all candidates with spatial overlap. 
To estimate the geometric term, pixels within the object mask $M_k^i$ are first back-projected into 3D using the depth map $D_k$. The spatial proximity is calculated as the 3D Intersection-over-Union (IoU) $S^{\textrm{geo}}_{i,j} = \mathrm{IoU}(B_j, p_{\text{project}})$, which is the overlapping ratio between the projected points and the existing bounding box $B_j$. The semantic term $S^{\textrm{emb}}_{i,j} = \cos(\mathbf{f}_k^i, \mathbf{f}_o^j)$ measures the cosine similarity between the new and existing object embeddings. 
The overall similarity integrates both cues through a weighted combination:
\begin{equation}
    S = \lambda S^{\textrm{geo}}_{i,j} + (1 - \lambda) S^{\textrm{emb}}_{i,j}.
\end{equation}
If no existing instance exceeds the similarity threshold $\delta_s$, a new object is initialized; otherwise, the matched instance is refined by updating its features as 
$\mathbf{f}_o^i = (n_o \mathbf{f}_o^i + \mathbf{f}_k^i) / (n_o + 1)$, 
where $n_o$ denotes the number of prior detections associated with object $o_j$, and by merging their corresponding Gaussians $\mathbf{g}_k \cup \mathbf{g}_j$. To enhance system efficiency and robustness, we select three key frames for each object to perform part-level decomposition and physical reasoning. Specifically, key frames are defined as the frames in which YOLO-World detections exhibit the largest mask areas while satisfying a high confidence threshold.

\subsection{Physical Reasoning} 
\begin{figure*}[t]
    \centering
    \includegraphics[width=.9\linewidth]{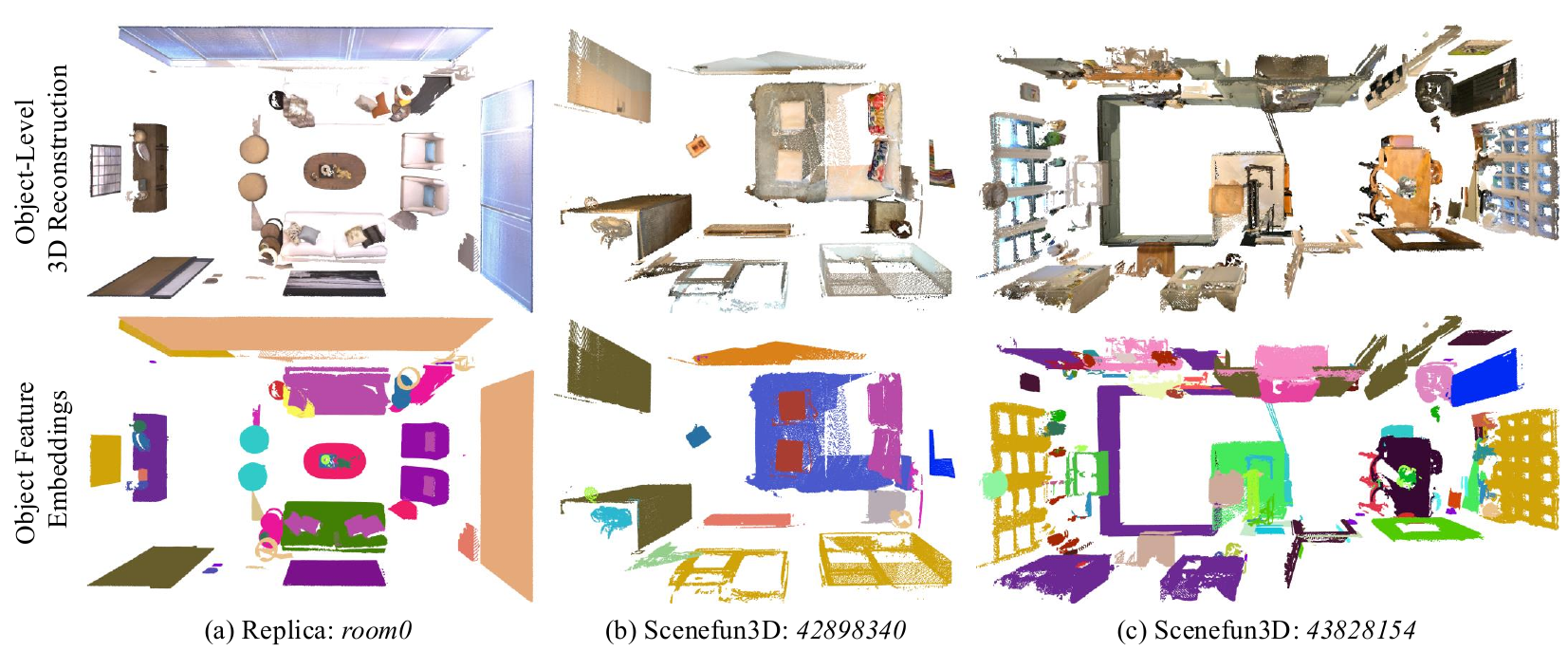}
    \vspace{-1mm}
    \caption{\textbf{Object Reconstruction Qualitative Results}. We present three reconstructed scenes from both a synthetic dataset (Replica) and real-world sequences (SceneFun3D). The first row shows the recognized object-level 3D reconstructions. The second row visualizes the corresponding object feature embeddings projected via PCA, where similar objects exhibit similar colors.}
    \label{fig:obj_res}
    \vspace{-2mm}
\end{figure*}

Understanding daily objects requires fine-grained estimation of part-level physics, as each part contributes various physical properties. The set of materials that exists in the world is extremely large and difficult to define. Driven by this, we call upon LLMs for visual-semantic reasoning about materials $P_j$ and their physical attributes, including Young’s modulus $\mathbf{E}$, Poisson’s ratio $\mathbf{\nu}$, and density $\mathbf{d}$. We then construct a material-aware field by propagating the inferred physical attributes to the corresponding 3D Gaussians.

\noindent \textbf{Articulation Estimation via Visual Prompting}
Reliable articulation estimation requires reasoning beyond closed-set datasets. To decompose each object into functional components, we apply MobileSAMv2~\cite{mobile_sam} to perform part-level segmentation on selected key frames, providing structured visual prompts for generalizable inference. In rigid-body kinematics, seven canonical lower-pair joint types exist (revolute, prismatic, cylindrical, helical, universal, spherical, planar). PhysGraph models two fundamental primitives: revolute and prismatic, which span rotational and translational degrees of freedom. This minimal yet expressive formulation provides a practical basis for representing a broad range of articulated motions. In principle, more complex joint types may be approximated as constrained compositions of these primitives (e.g., a screw joint couples rotation and translation along a shared axis).

\noindent\textbf{(i) Revolute joints.} Revolute joints connect two parts via hinges, with the rotation axis aligned along the hinge direction. Since hinges lie near part intersections, we extract vertices from part-level masks~\cite{mobile_sam} to form hinge candidates. These are indexed and visualized, and GPT-5 selects start and end vertices under the right-hand rule. The predicted 2D hinge and direction are lifted to 3D using camera intrinsics and poses. For cases where the hinge center is not on an edge (e.g., knobs), we assume an axis perpendicular to the supporting plane through the center or an annotated vertex. We further aggregate predictions across key frames and select the most geometrically consistent estimate to ensure physical plausibility.

\noindent\textbf{(ii) Prismatic joints.} Prismatic joints produce linear motion along a single axis, often partially occluded. As geometry alone may be ambiguous, we apply simple priors: inward/outward motions (e.g., drawers) translate perpendicular to the surface normal. For sliding motions, such as sliding windows, the potential
translation direction is constrained to two dimensions along the surface. This allows us to use the segmentation edges to prompt LLM to select the most appropriate direction.

\noindent \textbf{Physical property reasoning} Object-level and part-level segmentation provide detailed global to local visual prompts. Thus, during inference we not only query kinematic constraints but also ask LLM to return a dictionary of candidate materials regarding each part, e.g. $\{\text{Aluminum}:~E=69~\text{GPa},~\nu=0.33,~\rho=2700~\text{kg/m}^3\}$. Unlike existing works~\cite{zhai2024physical, shuai2025pugs}, we prompt the VLM with structured visual cues, including part-level masks, annotated hinge candidates, and part-isolated RGB appearances, which narrow the reasoning region and improve the accuracy of articulation and material estimation.

\begin{figure*}[t]
    \centering
    \includegraphics[width=\textwidth]{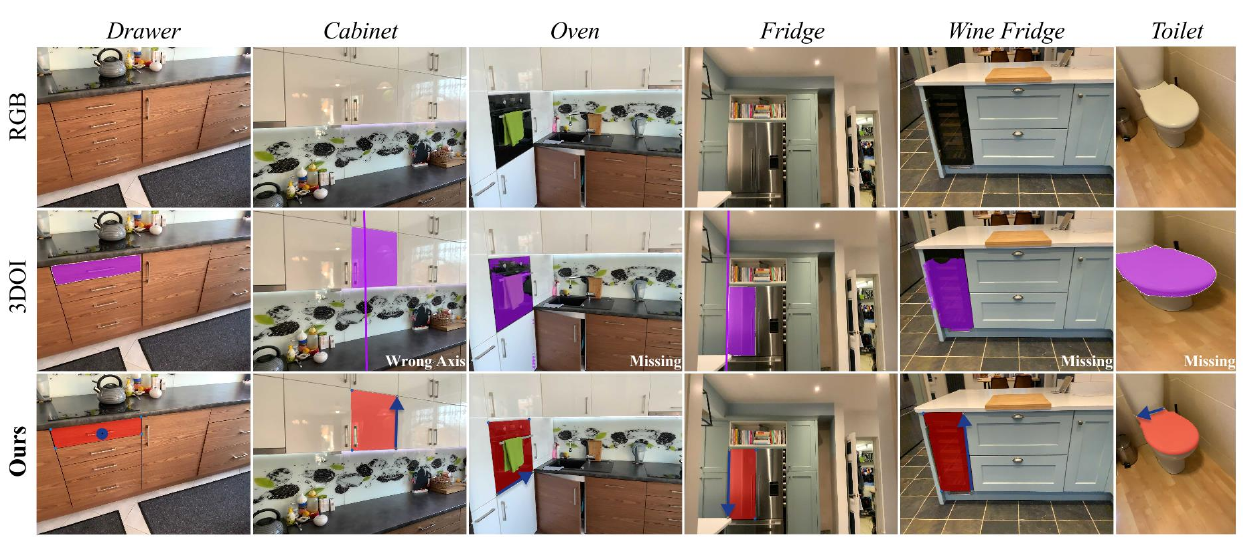}
    \caption{\textbf{Articulation Results}. Compared to 3DOI\cite{qian2023understanding}, PhysGraph generalizes across diverse cabinet, door, and toilet geometries, accurately identifying joint types and motion directions even in cluttered or unseen configurations, while baselines frequently fail or mispredict articulation. The visualization articulation axes follow the right-hand rule.}
    \label{fig:articulation_results}
    \vspace{-2mm}
\end{figure*}

\noindent \textbf{Adaptive Voxelization} To estimate object volume and mass from 3D Gaussians, we voxelize each object using an adaptive resolution determined by its bounding box. Because object sizes vary widely, a fixed voxel size would oversample large objects and miss details in small ones, leading to significant volume errors. Instead of a universal voxel size, we sample $N_v$ voxels inside each objects' 3D bounding box, where $N_v$ is manually chosen number to balance accuracy and efficiency. This adaptive strategy maintains geometric fidelity across scales while keeping computational cost consistent.

\noindent \textbf{Material Assignment} Assigning material candidate to each 3D Gaussian is challenging due to limited view points and complex object layout in real-world. To address this issue, we introduce an adaptive material propagation. We first estimate the visibility of each Gaussian primitive $s_j\in\mathbf{g}_i$ regarding each key frame, using input depth $D_k$. Then assign LLM proposed material to corresponding Gaussians. For Gaussians not covered in the visible view, we compute the cosine similarity between its learned visual feature $\phi_q$ and those of its $k$-nearest neighbors, and assign the most similar material prediction. We then project the per-Gaussian material probabilities into a volumetric field to enable physical estimation such as mass computation.

Specifically, let $\{v\}$ denote the voxels inside the object's AABB and $x_v \in \mathbb{R}^3$ be the center of voxel $v$. 
For each Gaussian primitive $s_j \in \mathbf{g}_i$ with centroid $\mathcal{X}_j$, covariance $\boldsymbol{\Sigma}_j$, opacity $\alpha_j$, and a one-hot material vector 
$\boldsymbol{\omega}(s_j) \in \mathbb{R}^C$ over $C$ material categories, we define its influence on voxel $v$ as
\begin{equation}
w_j(v) = \alpha_j \exp\Big(
-\tfrac{1}{2}\,(x_v - \mathcal{X}_j)^\top \boldsymbol{\Sigma}_j^{-1} (x_v - \mathcal{X}_j)
\Big),
\end{equation}
and normalize over all Gaussians that spatially overlap with $v$, denoted by $\mathcal{N}(v)$:
\begin{equation}
\tilde{w}_j(v) = 
\frac{w_j(v)}{\displaystyle \sum_{s_{j'} \in \mathcal{N}(v)} w_{j'}(v)}.
\end{equation}
The material probability of voxel $v$ is then obtained as a Gaussian-weighted aggregation of neighboring Gaussians:
\begin{equation}
\boldsymbol{\omega}(v) = 
\sum_{s_j \in \mathcal{N}(v)} \tilde{w}_j(v)\, \boldsymbol{\omega}(s_j),
\qquad \boldsymbol{\omega}(v) \in \mathbb{R}^C,
\end{equation}
and the final discrete material label is assigned by
\begin{equation}
m(v) = \arg\max_{c \in \{1,\dots,C\}} \,\omega_c(v).
\end{equation}
This material-aware voxel aggregation yields a dense, voxel-level material field that is consistent with the underlying 3D Gaussian representation and suitable for subsequent volume and mass estimation.

\subsection{Hierarchical Scene Graph Generation} 

Given the reconstructed set of 3D objects $\mathbf{v}_i \in \mathcal{V}_o$ obtained from the object detection module, we estimate their spatial relationships $\mathcal{E}_o$ to complete the object-level scene graph. 
We then incorporate results from the GVP module to generate part-level nodes $\mathcal{V}_p$ and functional relationships $\mathcal{E}_p$, forming a unified hierarchical scene graph that captures both structural and physical dependencies.

\noindent \textbf{Object-level scene graph.} 
The object-level scene graph focuses on spatial and semantic relationships among object instances. 
Each node $\mathbf{v}_i$ stores the object description, instance embedding $\mathbf{f}_i$, geometry $\mathbf{g}_i$, 3D bounding box $\mathbf{b}_i$, estimated mass $m_i$, and its associated part-level scene graph $\mathcal{G}_p$. 
Following \cite{gu2024conceptgraphs}, we construct a minimum spanning tree (MST) to identify potential edges among objects, providing a structural prior for scene organization. 
To determine semantic relationships, we prompt the VLM with object-annotated images and their corresponding 3D locations, obtaining relational descriptions such as “object $a$ attaches to object $b$” or “object $a$ is on object $b$.” 
This object-level graph serves as a foundation for high-level reasoning tasks such as spatial understanding, affordance prediction, and task planning.

\noindent \textbf{Part-level scene graph.} 
The part-level scene graph captures functionality and fine-grained physical attributes within each object. 
Each part node $\mathcal{V}_p$ encodes  articulation type, articulation position, and material or physical properties $\{\mathbf{E}, \boldsymbol{\nu}, \rho\}$. 
Edges $\mathcal{E}_p$ represent a functional description between connected parts, reflecting how individual components interact. 
This fine-grained graph enables downstream applications such as functionality reasoning, operation part retrieval, and physically grounded interaction simulation.

\begin{table}[t]
    \vspace*{2mm}
    \centering
    \caption{
        \textbf{Zero-shot semantic evaluations on Replica dataset~\cite{replica19arxiv}}. \colorbox{best}{Best}, \colorbox{second}{Second}, and \colorbox{third}{Third} best results are highlighted.
    }
    \small
    \setlength{\tabcolsep}{3pt}
    \begin{tabular}{lcccc}
        \toprule
        \textbf{Method} 
        & \textbf{mIoU$\uparrow$} 
        & \textbf{fIoU$\uparrow$} 
        & \textbf{mAcc$\uparrow$} 
        & \textbf{fAcc$\uparrow$} \\
        \midrule
        Open Gaussian~\cite{wu2024opengaussian}      & 6.82  & 15.41 & 16.66 & 18.08 \\
        Lang Splat~\cite{qin2023langsplat}         & 10.00 & 39.69 & 22.93 & 44.16 \\
        Grasp Splats~\cite{ji2024-graspsplats}       & 10.42 & 42.67 & 23.79 & 52.39 \\
        Omnimap~\cite{omnimap}            & \cellcolor{second}29.06 & \cellcolor{best}64.42 
                           & \cellcolor{second}44.14 & \cellcolor{best}72.22 \\
        Concept Fusion~\cite{conceptfusion}     & 4.75  & 25.30 & 19.29 & 28.99 \\
        Concept Graph~\cite{gu2024conceptgraphs}      & \cellcolor{third}16.46 & 35.69 
                           & 31.51 & 42.44 \\
        Open Fusion~\cite{10610193}        & 16.37 & \cellcolor{third}51.65 
                           & \cellcolor{third}35.15 & \cellcolor{third}60.37 \\
        Ours               & \cellcolor{best}32.74 & \cellcolor{second}60.51 
                           & \cellcolor{best}48.60 & \cellcolor{second}68.93 \\
        \bottomrule
    \end{tabular}
    
    \label{tab:segmentation}
\end{table}

\begin{table}[t]
\centering
\caption{
    \textbf{Articulation Estimation Evaluation.} 
}
\setlength{\tabcolsep}{3pt}
\renewcommand{\arraystretch}{1.1}

\resizebox{\columnwidth}{!}{%
    
    \begin{tabular}{l c c c c c}
    \toprule
    \textbf{Methods} 
    & \textbf{Total \#} % Header
    & \textbf{Pred \#} 
    & \textbf{Joint Acc.} %[\%] $\uparrow$ 
    & \textbf{Min. Dist.} %[cm] $\downarrow$ 
    & \textbf{Orient. Err.} %[$^\circ$] $\downarrow$ 
    \\
     
    &   % Header
    &  
    &   [\%] $\uparrow$ 
    &  [cm] $\downarrow$ 
    &  [$^\circ$] $\downarrow$ \\
    \midrule
    
    3DOI~\cite{qian2023understanding} 
    & \multirow{4}{*}{501} % Combine next 4 rows into one cell with value 264
    & 429 & 33.57 & - & - \\
    
    URDFormer~\cite{chen2024urdformer} 
    & % Empty cell here because of multirow
    & 244 & 55.33 & 61.46 & \textbf{1.86} \\
    
    DRAWER~\cite{xia2025drawerdigitalreconstructionarticulation} 
    & % Empty cell here
    & \textbf{441} & 68.25 & 18.28 & 2.07 \\
        
    \textbf{Ours} 
    & % Empty cell here
    & 429 & \textbf{96.04} & \textbf{6.33} & 5.84 \\
    \bottomrule
    \end{tabular}%
}
\label{tab:art_eval}
\vspace{-2mm}
\end{table}

\section{Experiments}

\begin{table}[t]
    \vspace*{2mm}
    \centering
        \caption{\textbf{Mass Estimation Comparison}. }
    \small
    \setlength{\tabcolsep}{3pt}
    \begin{tabular}{lcccc}
        \toprule
        Method & ADE~$\downarrow$ & ALDE~$\downarrow$ & APE~$\downarrow$ & MnRE~$\uparrow$ \\
        \midrule
        Image2mass~\cite{chen2024urdformer} & 21.07 &  1.919  & 4.787 & 0.156 \\
        GPT-5~\cite{xia2025drawerdigitalreconstructionarticulation}     & 25.49 & 1.896 & 5.657 & 0.150 \\
        \textbf{Ours} & \textbf{16.54} & \textbf{1.190} & \textbf{2.287} & \textbf{0.304} \\
        \bottomrule
    \end{tabular}
    
    \label{tab:mass_comparison}
\end{table}

We evaluate our approach at both the component and system levels. At the component level, PhysGraph achieves superior performance on semantic segmentation (Table~\ref{tab:segmentation}), articulation modeling (Table~\ref{tab:art_eval}), and per-object mass estimation (Table~\ref{tab:mass_comparison}), outperforming strong task-specific baselines in each setting. At the system level, PhysGraph constructs accurate 3D scene graphs (Table~\ref{tab:sg_eval}), enables task-driven affordance prediction (Table~\ref{tab:aff_eval}), produces physics-aware reconstructions (Fig.~\ref{fig:3DGS-to-MuJoCo}), and supports physically grounded robotic manipulation (Fig.~\ref{fig:down_manip}). 

Our pipeline employs CLIP ViT-L/14~\cite{ilharco_gabriel_2021_5143773} and DINOv3 ViT-B/16~\cite{siméoni2025dinov3} to extract object-level feature embeddings, followed by GPT\textendash5~\cite{openai2025gpt5} for visual reasoning. We set $\lambda=0.3$ for the weighted similarity computation and use a threshold of $\delta_s=0.25$. The voxel count is fixed to $N_v=64$, providing a balanced tradeoff between accuracy and computational efficiency. All experiments are conducted on a desktop with an Intel Ultra 9 CPU and an NVIDIA RTX~5090 GPU (32\,GB). For a typical scene containing 500 frames, PhysGraph requires approximately 25 minutes to full process. This includes 15 minutes for object-centric perception, 1 minute for part-level segmentation, and 9 minutes for physical reasoning, including GPT inference (8 minutes) and material assignment (1 minute).

\subsection{3D Semantic Segmentation}

We first evaluate the quality of the reconstructed 3D map using an open-vocabulary 3D segmentation task that retrieves semantic labels for reconstructed 3D regions. Specifically, we compare PhysGraph against open-vocabulary baselines on zero-shot semantic segmentation, including 3DGS-based methods~\cite{wu2024opengaussian, qin2023langsplat, ji2024-graspsplats, omnimap} and scene-graph approaches~\cite{conceptfusion, gu2024conceptgraphs, 10610193}. Using dataset labels, we compute similarities between each instance’s fused feature and all category embeddings, assigning its 3D Gaussians to the most similar class. Table~\ref{tab:segmentation} reports results on Replica~\cite{replica19arxiv} using mIoU, fIoU, mAcc, and fAcc. PhysGraph achieves the highest mIoU and competitive performance across other metrics, surpassing both 3DGS- and scene-graph-based baselines. These gains arise from our object-centric perception module (Section~\ref{p:o_perception}), which enforces cross-view semantic consistency during object association and yields more stable, discriminative representations.

\subsection{Physical Property Estimation}

We compute object-level mass using the generated material field. However, to our knowledge, no existing dataset provides ground-truth mass annotations of multiple objects at the scene level. We therefore use Behavior-1K~\cite{li2024behavior1k} to collect RGB-D sequences in two scenes, \texttt{Rs\_int:living\_room\_0} and \texttt{Wainscott\_0:dining\_room}, covering 25 objects from small appliances to furniture, and manually obtain their ground-truth masses.

We attempted to adapt prior methods~\cite{shuai2025pugs, zhai2024physical} by masking objects and providing camera extrinsics, but they fail under missing global scale and sparse views. Instead, we compare with Image2mass~\cite{pmlr-v78-standley17a} and GPT-5~\cite{openai2025gpt5}, both operating on 2D images. Image2mass predicts mass from segmented object images, while GPT-5 is prompted with the same visual input as PhysGraph to output a numerical estimate.

Following~\cite{ zhai2024physical}, we report four metrics: Absolute Difference Error (ADE), Absolute Log Difference Error (ALDE), Absolute Percentage Error (APE), and Minimum Ratio Error (MnRE). As shown in Table~\ref{tab:mass_comparison}, PhysGraph outperforms both baselines across all metrics, yielding lower and more stable errors. This improvement arises from addressing a key limitation of 2D methods, the lack of reliable volume estimation, by leveraging part-level material reasoning and adaptive voxelization that preserve geometric details.

\subsection{Articulation Modeling}

\begin{table}[t]
\centering
\caption{
    \textbf{Scene graph evaluation.}
    \colorbox{best}{Best}, \colorbox{second}{Second}, and \colorbox{third}{Third} best results are highlighted.
}
\setlength{\tabcolsep}{2pt} % Optimized spacing

\resizebox{\columnwidth}{!}{%
    \begin{tabular}{l cc cc cc cc cc}
    \toprule
    & \multicolumn{2}{c}{Obj.}
    & \multicolumn{2}{c}{Fun.}
    & \multicolumn{2}{c}{All Nodes}   % <--- Swapped: Nodes is now here
    & \multicolumn{2}{c}{Edge}    % <--- Swapped: Edge is now here
    & \multicolumn{2}{c}{Trip.}
    \\
    \cmidrule(lr){2-3} \cmidrule(lr){4-5} \cmidrule(lr){6-7}
    \cmidrule(lr){8-9} \cmidrule(lr){10-11}

    \textbf{Methods}
    & R3 & R10 
    & R3 & R10 
    & R3 & R10   % <--- Nodes metrics
    & R5 & R10   % <--- Edge metrics
    & R5 & R10
    \\
    \midrule

    FG~\cite{rotondi2025fungraph}
    & 40.8 & 64.3 
    & 33.7 & 51.9 
    & 36.1 & 56.1           % Nodes Data
    & 10.5 & 15.2           % Edge Data
    & 1.03 & 2.56 \\

    OFG~\cite{zhang2025open}
    & \Best{81.8} & \Best{87.8} 
    & \Sec{71.0} & \Sec{79.5} 
    & \Sec{73.0} & \Sec{82.8}   % Nodes Data
    & \Sec{88.1} & \Sec{96.2}   % Edge Data
    & \Sec{60.4} & \Sec{70.3} \\

    CG ~\cite{gu2024conceptgraphs}
    & \Thi{68.4} & \Thi{75.3} 
    & 4.20 & 6.41 
    & 21.3 & 28.1             % Nodes Data
    & \Thi{72.2} & \Thi{82.8}   % Edge Data
    & 2.11 & 5.41 \\

    O3D~\cite{koch2024open3dsg} 
    & 55.5 & 64.7 
    & \Thi{51.9} & \Thi{59.7} 
    & \Thi{53.2} & \Thi{62.8}   % Nodes Data
    & 70.1 & 80.2             % Edge Data
    & \Thi{34.8} & \Thi{45.2} \\

    \textbf{Ours}
    & \Sec{73.1} & \Sec{81.5} 
    & \Best{82.2} & \Best{88.2} 
    & \Best{78.8} & \Best{83.2} % Nodes Data
    & \Best{91.2} & \Best{97.5} % Edge Data
    & \Best{65.8} & \Best{73.2} \\
    \bottomrule
    \end{tabular}%
}
\label{tab:sg_eval}
\vspace{-2mm}
\end{table}

The performance of articulation estimation relies on both perception (i.e., identifying articulation location) and reasoning (i.e., estimating articulation type). To validate the effectiveness of our method, we assess both parts in Table~\ref{tab:art_eval}. Specifically, we compare our method with 3DOI~\cite{qian2023understanding}, USDFormer~\cite{chen2024urdformer}, and DRAWER~\cite{xia2025drawerdigitalreconstructionarticulation}. For fairness, all key frames are provided to the 2D baselines. Evaluation is conducted on 30 articulation-rich scenes from SceneFun3D~\cite{delitzas2024scenefun3d}. Following the MultiScan protocol~\cite{NEURIPS2022_3b3a83a5}, we report the number of articulated objects, joint-type accuracy, and hinge metrics: Minimum Distance (MD) between predicted and ground-truth joint lines, and Orientation Error (OE) between joint axes. For prismatic joints, we report only OE due to origin invariance; for revolute joints, both MD and OE are measured. As shown in~\autoref{tab:art_eval}, PhysGraph achieves over 96\% joint-type accuracy, significantly outperforming all baselines. It also attains the lowest MD among 3D methods while maintaining competitive OE, demonstrating the benefit of object-centric reasoning and part-level visual prompting for robust articulation estimation.

\subsection{Scene Graph Generation}

\begin{table}[t]
\vspace*{2mm}
\centering
\caption{
    \textbf{Affordance Success Rate.} 
}
\small 
\setlength{\tabcolsep}{6pt} 
\renewcommand{\arraystretch}{1.1}

\begin{tabular}{lcc} % Changed from 4 columns to 3 columns: Method, Category, Success
\toprule
\textbf{Methods} & \textbf{Task} & \textbf{Success Rate} [\%] \\
\midrule

% --- Group 1: 3D Segmentation ---
Fun3DU~\cite{corsetti2025fun3du}       & \multirow{2}{*}{3D Segmentation}     & 33.11 \\
Search3D~\cite{takmaz2025search3d}     &                              & 26.96 \\
\cmidrule{1-3} % Adjusted separator line for 3 columns

% --- Group 2: Scene Graph ---
ConceptGraph~\cite{gu2024conceptgraphs} & \multirow{5}{*}{Scene Graph} & 4.44 \\
Open3DSG~\cite{koch2024open3dsg}     &                              & 11.98 \\
FunGraph~\cite{rotondi2025fungraph}   &                              & 37.03 \\
OpenFunGraph~\cite{zhang2025open} &                              & 43.68 \\
\textbf{Ours} &                              & \textbf{50.68} \\

\bottomrule
\end{tabular}
\label{tab:aff_eval}
\vspace{-2mm}
\end{table}

We evaluate scene graph generation in terms of node detection and relation estimation. Experiments are conducted on the curated SceneFun3D dataset~\cite{rotondi2025fungraph}, which provides ground-truth nodes and relationships for 20 indoor scenes. We compare PhysGraph with four representative baselines~\cite{rotondi2025fungraph, zhang2025open, gu2024conceptgraphs, koch2024open3dsg}. Following~\cite{zhang2025open}, we report Top-$K$ Recall@K. For node-level evaluation, we measure object recall, functional element recall, and overall node recall. For relationship evaluation, we report edge recall and triplet recall. As shown in Table~\ref{tab:sg_eval}, PhysGraph outperforms most baselines across metrics, particularly on functional elements. We attribute this improvement to our part-level reasoning mechanism, which decomposes object instances and captures fine-grained structural elements.

\subsection{Task Affordance Prediction}

\begin{figure}[t]
    \vspace*{2mm}
    \centering
    \includegraphics[width=.9\linewidth]{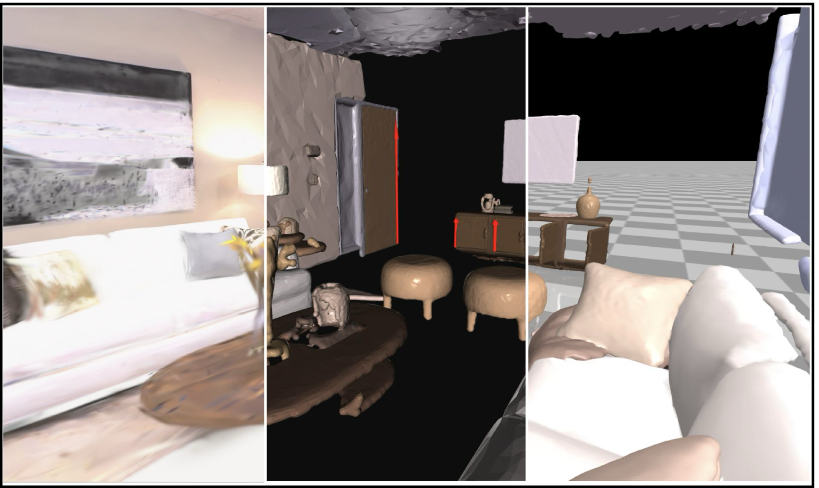}
    \caption{Reconstruction to Simulation: the \texttt{room\_0} scene from the Replica dataset. From left to right: 3DGS reconstruction, watertight mesh with movable revolute joint locations and directions encoded in the red arrows, and the resulting simulation environment in MuJoCo.}
    \label{fig:3DGS-to-MuJoCo}
    \vspace{-2mm}
\end{figure}

In the following experiments, we further demonstrate the performance of our task-driven affordance grounding in SceneFun3D~\cite{delitzas2024scenefun3d} focusing on functional interactive parts. We mainly compare our method with 3D segmentation~\cite{corsetti2025fun3du, takmaz2025search3d} and related scene graph baselines~\cite{rotondi2025fungraph, zhang2025open, gu2024conceptgraphs, koch2024open3dsg}. Specifically, each query consists of a text-based description of the task and the goal is to retrieve the functional interactive elements, for example "Open the bottom left window above the radiator". To achieve this, we convert our hierarchical scene graph into a JSON format and then call upon GPT-5 to find the Part ID(s) that solve the query. For evaluation, we retrieve ground truth point clouds and our Gaussians to compute the 3D point cloud intersection over union (IoU). We count a query as passed if the IoU is at least 25\%.

\subsection{Downstream Applications}

\noindent \textbf{Physics-aware Reconstruction} As shown in Fig.~\ref{fig:3DGS-to-MuJoCo}, a key advantage of our framework is generating physics-aware scenes for grounded robotic interaction. 
We convert the hierarchical 3D scene graph $\mathcal{G}_o=(\mathcal{V}_o,\mathcal{E}_o)$ into a MuJoCo format~\cite{todorov2012mujoco}. 
Each object node $\mathbf{v}_i=(\mathbf{g}_i,\mathbf{p}_i,\mathcal{G}_p)$—including geometry, articulation, and physical attributes—is reconstructed as a watertight mesh via Marching Cubes on its 3DGS field. The meshes preserve real-world scale and orientation, and material parameters $\{\mathbf{E}, \nu, \rho\}$ are mapped to corresponding MuJoCo elastic and damping coefficients. Articulations are instantiated as revolute or prismatic joints, producing a fully parameterized MuJoCo XML file. 

\begin{figure}[t]
    \vspace*{2mm}
    \centering
    \includegraphics[width=.9\linewidth]{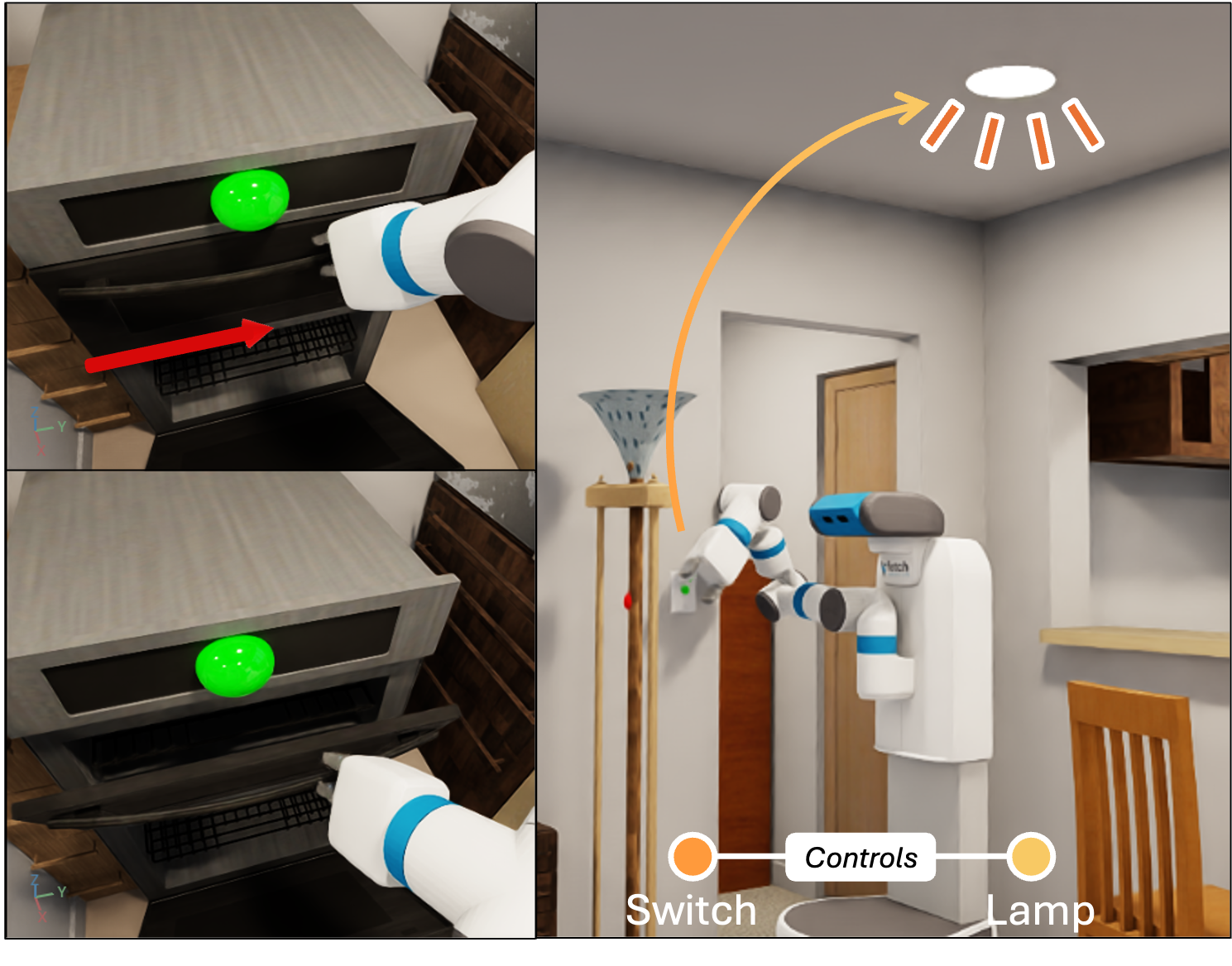}
    \caption{\textbf{PhysGraph for manipulation}. A robot executes manipulation tasks guided by the structured scene graph. \textit{Left:} the robot localizes the oven handle and infers the correct articulation axis to open the door. \textit{Right:} the robot reasons over functional relationships to interact with the environment.}
    \label{fig:down_manip}
    \vspace{-2mm}
\end{figure}

\noindent \textbf{Robotic Manipulation} The generated 3D scene graph can also support robotic manipulation for task queries. Shown in Fig.~\ref{fig:down_manip}, users use the converted JSON file to locate the interactive part and infer articulation hinge positions. We test the results in Behavior-1K by sending 3D position of an interactive part and the robot then navigates to and interacts with the part.

\section{Conclusion}

In this work, we introduced PhysGraph, a physics-aware 3D scene representation that addresses key limitations in scene-level physical grounding. Our experiments demonstrate that a 3D scene graph serves as a natural and effective bridge between the high-level reasoning capabilities of large language models and fine-grained spatial understanding. This design enables PhysGraph to infer kinematic and material constraints, support accurate articulation and mass estimation, and deliver strong affordance predictions for downstream tasks. We believe this simple yet effective framework provides a foundation for integrating physical reasoning into robotic systems.

% However, our pipeline assumes known camera poses for reliable object-centric perception, and its performance is influenced by the quality of upstream 3D reconstruction. These dependencies may become bottlenecks in dynamic or cluttered environments. 
% Looking ahead, \textbf{PhysGraph} offers a compact, interpretable, and physically grounded scene representation that can function as persistent spatial memory for embodied agents. Its generalizable structure is well-suited for downstream applications such as long-horizon manipulation, interactive planning, and closed-loop policy learning. 

\bibliographystyle{IEEEtran} 
\bibliography{root}
\end{document}